# A LOW-COST WAVE/SOLAR POWERED UNMANNED SURFACE VEHICLE

Moustafa Elkolali
*Department of Mechanical, Electronic and Chemical Engineering*
*Oslo Metropolitan University*
Oslo, Norway
elkolali@oslomet.no

Ahmed Al-Tawil
*Department of Mechanical, Electronic and Chemical Engineering*
*Oslo Metropolitan University*
Oslo, Norway
ahmedalt@oslomet.no

Lennard Much
*Department of Mechanical Engineering*
*University of Applied Sciences Kiel*
Kiel, Germany
lennard.much@student.fh-kiel.de

Ryan Schrader
*Department of Mechanical Engineering*
*Michigan Technological University*
Michigan, USA
rmschrad@mtu.edu

Olivier Masset
*Department of Mechanical Engineering*
*Ecole Nationale d'Ingénieurs de Tarbes*
Tarbes, France
Olivier.masset@enit.fr

Marina Sayols
*Electronic Engineering Department, Universitat Politècnica de Catalunya*
Barcelona, Spain
Marina.sayols@estudiant.upc.edu

Andrew Jenkins
*Department of Mechanical Engineering, Glasgow Caledonian University*
Scotland, UK
eajenki203@caledonian.ac.uk

Sara Alonso
*Department of Industrial Design Engineering*
*Universidad Politécnica de Madrid* Madrid, Spain
sara.alonso.sanz@alumnos.es

Alfredo Carella
*Department of Mechanical, Electronic and Chemical Engineering*
*Oslo Metropolitan University*
Oslo, Norway
alfcar@oslomet.no

Alex Alcocer
*Department of Mechanical, Electronic and Chemical Engineering*
*Oslo Metropolitan University*
Oslo, Norway
alepen@oslomet.no

*Abstract*— This paper presents a prototype of a low-cost Unmanned Surface Vehicle (USV) that is operated by wave and solar energy which can be used to minimize the cost of ocean data collection. The current prototype is a compact USV, with a length of 1.2m that can be deployed and recovered by two persons. The design includes an electrically operated winch that can be used to retract and lower the underwater unit. Several elements of the design make use of additive manufacturing and inexpensive materials. The vehicle can be controlled using radio frequency (RF) and a satellite communication, through a custom developed web application. Both the surface and underwater units were optimized with regard to drag, lift, weight, and price by using recommendation of previous research work and advanced materials. The USV could be used in water condition monitoring by measuring several parameters, such as dissolved oxygen, salinity, temperature, and pH.

## I. INTRODUCTION

Unmanned Surface Vehicles (USVs) are steadily becoming common platforms and replacing/complementing manned vessels in many of the environmental monitoring and ocean data gathering operations [1]. Traditional types of research and ocean monitoring systems require fuel or electricity, but there is a large potential for systems that are fully energy self-sufficient. Various designs for a new type of autonomous platform have been created to meet future requirements [2]. A special class of these vehicles is suitable for long-range/duration operations as they use waves, wind, and solar power for propulsion or electric power source, but they still require a high operational cost.

Currently, various designs of USVs with multiple different applications have been developed in the past decade. Many of the existing models still have limited autonomy and still rely on experimental platforms and remote support [3]. The vehicles are powered by electrical, fossil, wind, solar, or wave energy and differ in size, shape, and navigation systems.

"Liquid Robotics" has pioneered the field with the development of their 'Wave Glider' models. With "Wave Glider SV2" and "Wave Glider SV3". The company developed an autonomous USV which is exclusively powered by wave and solar energy.

"Wave Glider SV3" consists of a monohulled float connected to a submerged glider and contains various sensors. The unit can operate self-sufficiently for up to one year. The disadvantages of these models, however, are their size and weight, which turns the recovery of these vehicles into a challenging operation.

Besides "Liquid Robotics", "AutoNaut" is another company that develops wave-powered USVs. In contrast to the 'Wave Glider', the "AutoNaut 3.5" and "AutoNaut 5.0" models consist of a single floating unit with hydrofoils located just below the surface for propulsion. Although these systems are relatively compact, the overall size and weight represent a challenge in deployment and recovery.

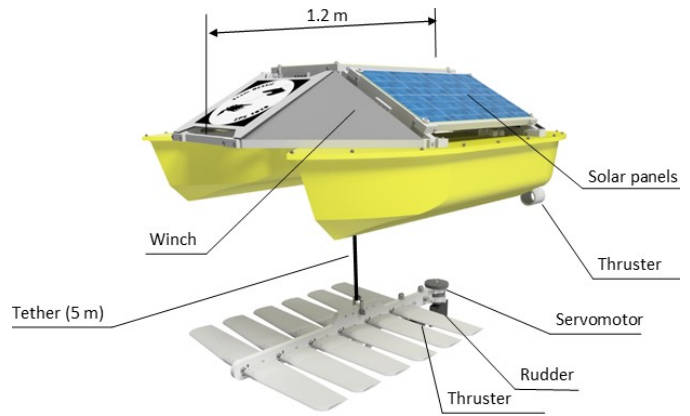

Fig. 1. 3D design of the proposed USV and its components

The existing solutions have high costs and are oriented towards open ocean long-range operations. There is a need for lower-cost solutions that can be used in gulfs, fjords, and coastal areas. The proposed USV has been designed to be operated by both wave and solar energy, with considerations for deployment and recovery at a fraction of the cost of current systems and will be desirable to a wider number of users. The 3D design of the USV is shown in Fig. 1. Costs and weight have been minimized throughout the design to appeal to a larger user base. This system allows for a variety of different purposes within the governmental, environmental, and research industries. By optimizing size, weight, and cost as well as an improved recovery and deployment stages, the collection of data from the oceans will be simplified.

## II. Surface Unit

The vehicle is composed of two main sections, a surface, and a submerged unit, linked together electrically and mechanically a tether. The surface unit is a vital part of the USV design. It is used to store electronic components, solar panels, batteries, and essential communication equipment.

### A. Hulls

USVs typically consist of a single hull filled with sensors, batteries, and navigation components and are designed to be utilized in the ocean. The hull geometry and style are important features in the design of a USV, as it can greatly affect the efficiency and stability whilst deployed. There are two styles of hull design, monohull, and multihull. Monohulls are commonly used in most vessels as they are easy to construct and can be easily scaled, multihulls are less common.

Multihull designs have huge improvements in stability over monohull designs concerning the roll and wave of the vessel [4]. This increased stability is a result of multihulls typically having a wider footprint than monohulls. This design benefits a USV as there is more space inside of the hull that can be used to fit more data-collecting sensors and more batteries to power the electronics [5]. Another effect of choosing a multihull concept over a monohull is increased flexibility in mounting and securing components, due to the addition of another hull. Having more free space allows for the addition of ballast to be fitted, enabling the center of weight to be redistributed for a more optimal and stable design [6]. Another advantage of using a multihull design is the creation of more space for solar panels to be mounted on a frame, increasing the solar power harvested.

A multihull design was selected to be used as the float for the surface unit. To maximize efficiency and reduce drag, the optimal distance between the two hulls must be determined. The separation distance was gathered by obtaining the minimum drag coefficient. According to [7] separation distance of half the length of the hulls was the most effective and produced the least amount of drag.

As mentioned before, the main objective of the design was to obtain a smaller and lighter USV. A hull length of 1.2 meters was determined to be the desired size as it is large enough to place the solar panels and the electronic components. This dimension is less than half the length of "Wave Glider SV3" which is 3.05 meters in length [8].

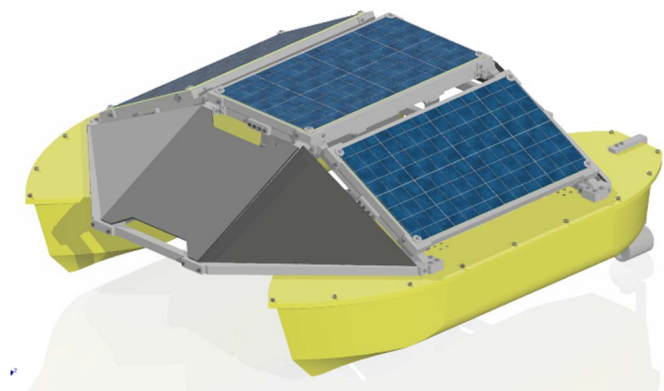

Fig. 2. 3D design of the Surface unit

Together, the two hulls can support up to 75kg. Fig. 2 shows the 3D design of the two hulls assembled with the rest of the Surface unit components.

The hull is to be manufactured using fiberglass. The material represents the best solution given its lightweight, low cost, and the possibility of manufacturing complex shapes. The manufacturing process, vacuum infusion, requires resin and fiberglass, which acts as a reinforcement, and consists of applying layers of both materials until the desired strength and thickness is obtained. The process is done by making a single mold, which will be used in manufacturing the two hulls. Fig. 3 shows the steps of manufacturing the mold.

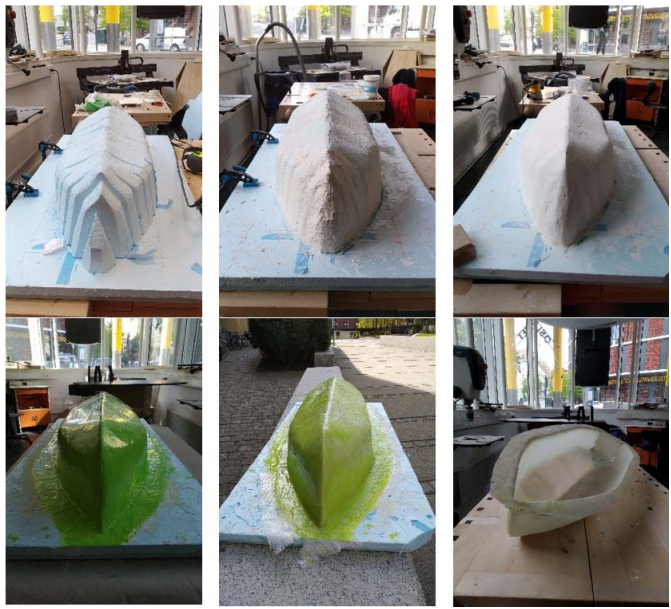

Fig. 3. Manufacturing steps of the hull from fiber glass

The lids that cover and seal the two hulls were also manufactured from fiberglass, by using vacuum infusion as shown in Fig. 4.

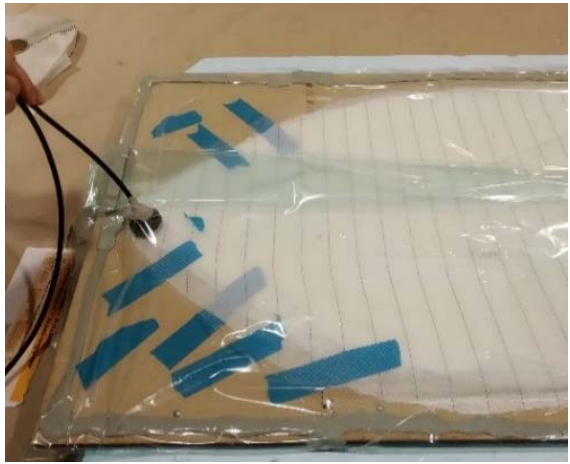

Fig. 4. Hull lid manufactured from fiber glass by vacuum infusion technique

*B. Frame*

Choosing a multihull design created the requirement of designing a frame. The frame connects the two hulls and ensures the desired separation distance. It is also used to mount the solar panels and winch assembly. To meet these requirements, the frame had to be structurally strong to support the weight of the Wave Glider along with the other components. The frame has also been designed to be modular as it is assembled with bolts. Fig. 5 shows the design of the structural frame.

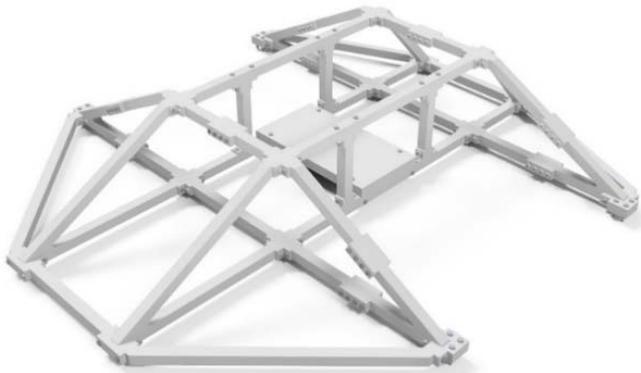

Fig. 5. Frame assembly 3D design

To connect both hulls, a strong and light material was required. Polyvinyl chloride (PVC) material was chosen for the frame due to its machinability, mechanical properties, and weight. The frame was manufactured using a 3D CNC router as shown in Fig. 6.

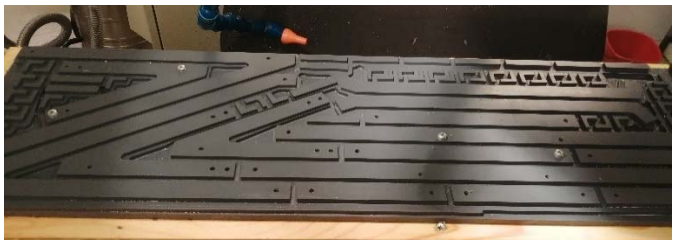

Fig. 6. Frame during manufacturing process

The frame is then assembled all together using stainless steel bolts and locknuts to prevent rust and to ensure the durability of the design. Fig. 7 shows the assembled frame, before attaching the solar panels and the rest of the components.

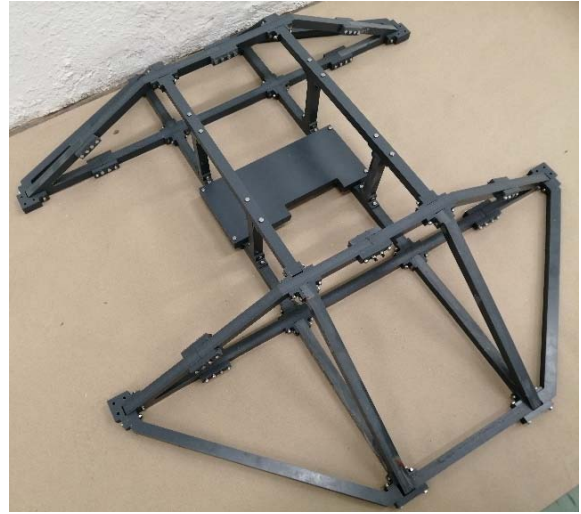

Fig. 7. Assembled modular frame connecting the two hulls

The frame is covered from the top by three solar panels. All three solar panels are waterproof from all sides and covered by a protective anti-scratch transparent layer from the top. A modular frame was created to mount each solar panel. This enables the panels to be easily mounted to the main frame between the hulls. Fig. 8 shows the solar panels mounted on the frame of the surface unit.

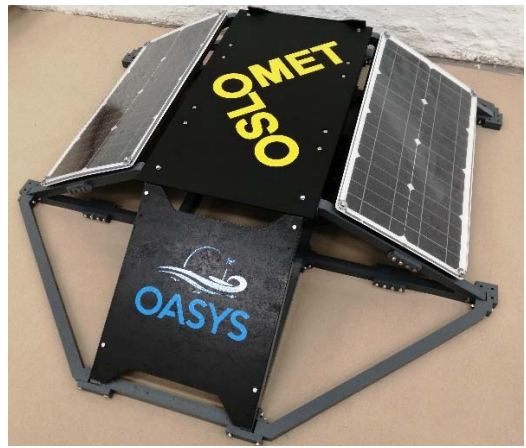

Fig. 8. Solar panels mounted on the frame of the surface unit

*C. Winch*

To reach the desired design specifications, a winch was required to retract and extend the tether so that two able-bodied persons can deploy and recover the system. Existing 'Wave Gliders' need a crane for deployment and recovery as they are unable to retract the several meters tether. The crane is used to lift the entire USV out of the water which tends to be costly and challenging. By utilizing a winch built into the surface unit, the 5-meter tether can be retracted, bringing the underwater unit up to the surface. This enables the whole USV to be lifted out of the water simultaneously, simplifying the recovery process.

The winch is mounted to the frame and located between the two hulls as to evenly distribute the weight across the boat. It is secured to the base of the frame with bolts. The winch had several design requirements that influenced the final decision. It needed to be small, lightweight, and have a DC input. Additionally, the winch cable had to be strong enough to support the weight of the underwater unit. Lastly, the winch had to be able to hold its position without being powered as to not drain the energy stored in the batteries. The used winch is rated at 1.8kW at full capacity, with the ability to lift 900kg. This is more than enough to retract the tether, overcome drag, and lift the submerged unit to the surface unit. The winch is rated IP68, which means that it is sealed against continuous submersion in water. Fig. 9 shows a picture of the winch before it was assembled to the frame.

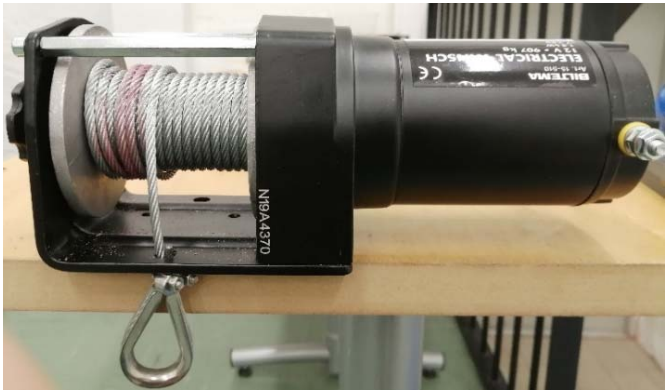

Fig. 9. IP68 rated winch for retracting and deploying submerged unit.

## III. UNDERWATER UNIT

Resting below the water level, the underwater unit provides the propulsion for the entire system by using wave energy. The connection with the surface unit is achieved by a 5-meter tether. This length allows the underwater unit to operate in calmer conditions, where its movement is not affected by the vortices near the surface. Depending on the state of the waves, the six oscillating wings that are mounted on the frame are set into motion and produce forward thrust. Furthermore, a thruster, a rudder, and waterproof housing for the servomotor are installed. The thruster is only used in low sea states, high opposing currents, or in cases of emergency. The whole assembly of the submerged unit can be seen in Fig. 10.

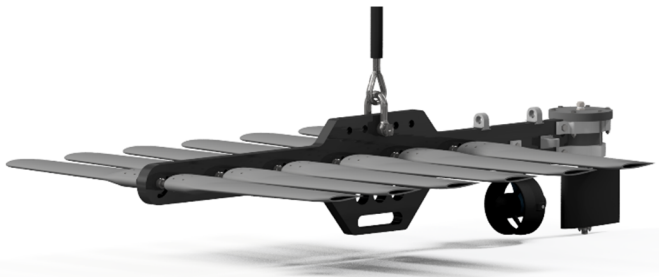

Fig. 10. Assembly of the underwater unit

### A. Frame

The underwater unit has been designed with reference to current systems and research while implementing new design considerations. The main objective was to achieve the highest possible functionality regardless of the reduction in size, weight, and cost. All major components for the propulsion and steering of the underwater unit are mounted to the frame. To achieve the goal of saving as much weight as possible and in consideration of material and manufacturing costs, it was decided to use polymers instead of light metals such as aluminum. The polymers have high machinability and tensile strength at a low price. Density, tensile strength, and price were the decisive factors for comparing Polyoxymethylene (POM), Polyvinyl chloride (PVC), and Polyamide-Nylon (PA) materials. Eventually, PVC was chosen as the material for the frame. The frame dimensions are only 950mm x 185mm x 20 mm (LxWxH).

### B. Wings

The underwater unit is powered by oscillating wings. The design of the wings must consider various parameters, such as the profile, chord length, or length of each wing along with the spacing between them. To select the most efficient wing profile, various aerofoil shapes defined by the National Advisory Committee for Aeronautics (NACA) were taken into consideration. Depending on the NACA series and geometry, the wings differ in parameters such as thickness and camber. There are many different approaches when it comes to selecting the correct wing profile. Reference [9] compared a simple plate profile with a symmetrical NACA-0006 profile and concluded that the symmetrical NACA profile produces a significantly higher forward lift. Reference [10], however, made a comparison between a symmetrical NACA-0012 profile and a NACA-2315 cambered wing profile which showed that the maximum lift-to-drag ratio of the symmetrical profile is higher. The symmetrical profile NACA-0012 was chosen for the wing cross-section since it has better hydrodynamic properties and provides a maximum thrust throughout the allowed rotation angle [11].

In addition to the profile, the wing is also defined by the chord length and wingspan. The chord length is the length from the nose of the wing to the trailing edge. To minimize the overall size, while ensuring a reliable mounting to the frame, a chord length of 0.12 m was chosen. This size was based on the smaller lengths used in previous experiments in conjunction with a NACA-0012 wing profile conducted by [12], but slightly adjusted for a larger mounting area. The wingspan was chosen to be 0.325 m, compared to the "Wave Glider SV2", which has a wingspan of 0.55 m [13]. This length still provides sufficient stability of the whole unit, while reducing the overall size.

In order to determine the number of sets of wings are the most efficient for providing forward propulsion, [12] tested different numbers of wings under the same conditions. The wing profile used was a NACA-0012 with a wingspan of 0.2 m and a chord length of 0.1 m. The results showed that the thrust coefficient, which is an indication of the thrust generated by the flapping wings, increases proportionally with the number of wings, which is explained by the increase in interference strength between the foils. The vortices created by the first wing also affects the following ones negatively. The study also shows that the average thrust coefficient plateaus as the number of wings increase, however with a higher number of wings, the thrust coefficient is leveling off. As a result, it was decided to use the best number of sets of wings for the current design, which is six sets.

The distance between each of these six sets of wings was also considered. In another experiment, [12] have investigated the influence of different spacings between the wings and how it affects thrust. In the experiment, the chord length used on a NACA-0012 profile was 0.1 m. The results show that the thrust coefficient and distance between the wings are inversely proportional. At a reduced distance, the interferences of the vortices between the wings are stronger, which leads to a higher thrust coefficient. Based on this phenomenon, a spacing of 0.1 of the chord length (0.1c) was chosen, which corresponds to 12 mm for the chosen chord length of 120 mm.

An important factor for the forward propulsion of the system is the determination of the optimal limit angle of the wings. The limit angle determines by how many degrees the wings of the submerged unit can oscillate. The use of different limit angles can lead to different propulsive and hydrodynamic performances. Reference [14] conducted a numerical simulation for five different limit angles of ±10°, 15°, 20°, 25°, and 30° were tested. It was discovered that at a limit angle of ±20°, the thrust coefficient reaches its maximum. If the angle exceeds ±25° the foil begins to stall, which has a negative impact on the thrust coefficient. As a result, ±20° was selected as the maximum limit angle.

In practice, however, this angle can only be realized for the unit's falling movement. During the rising motion, an enormous force can act on the surface area of the wings, depending on the sea state, so they could experience fatigue in rough sea states. As a result, a relief angle of -90° has been implemented into the design to avoid the risk of the break of the wings or the supporting shafts. With this consideration, realized from "Liquid Robotics SV3" model, the overall limit angle was increased to 110° which is restricted by a slot that has been cut into the frame of the glider, displayed in Fig. 11. The tension of the springs allows the wings to continue with an optimum hydrodynamic performance at ± 20°, but still allows the wings to become completely vertical while being pulled up at a high speed.

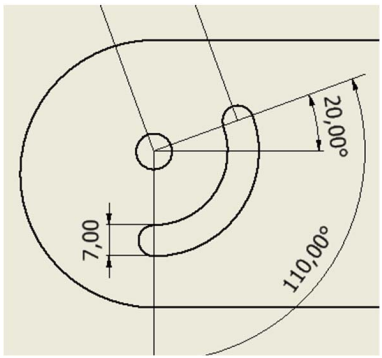

Fig. 11. Slot for limiting shaft of the wing.

Before the wings were mounted to the actual frame, a PVC model was used to test their functionality. The wings were mounted to the limiting shaft and pivoting shaft with four set screws as shown in Fig. 12.

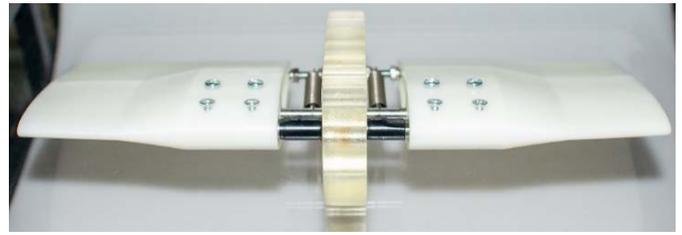

Fig. 12. Dummy test section of wing mounting

*C. Springs and Shafts*

Two types of springs were concluded to be possible fits for the system. Both the use of torsion springs and extension springs can be found in existing research. At first, torsion springs were selected for the design, but due to uncertainties in assembly and a relatively high price, it was decided to use extension springs instead. The extension springs are installed between the frame and the wings. One eye of the spring is connected to the limiting shaft and the other eye is attached to a small shaft protruding from the frame, as illustrated in Fig. 13. During the oscillation process, they ensure that the wings are returned to their neutral position after they have reached the maximum angle of rotation.

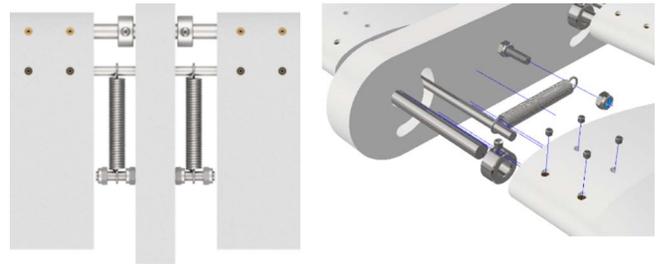

Fig. 13. 3D design of the linear spring mounting

To achieve the optimal thrust performance, the choice of the right spring stiffness is an important factor because it controls the pitching of the wings. If the selected stiffness is too high, the wing might not deflect sufficiently and therefore lead to poorer performance. Alternatively, choosing a stiffness that is too low could lead to a non-optimal angle of attack and have a negative effect as well [15]. To get an indication for the choice of the spring stiffness, it was orientated towards the values used by [11] during their experiments to analyze the dynamic performance using CFD software. They used a spring rate of 2.9 N/mm with a simplified wing surface area of 0.096 m². The ideal spring rate was chosen based on the ratio 0.406 between the simplified wing surface areas, given that the prototype's wing is 0.039 m². This ideal rate was determined to be 1.178 N/mm.

*D. Rudder*

The rudder is the only actuated mechanism on the whole vessel and is vital to follow the correct heading. The rudder size considerations were first modeled after reference [13] study, where a height of 130 mm and 210 mm chord length is utilized. The sizing was decreased to fit proportionally with the submersed unit's smaller size. After alterations, the prototype version has the parameters of 90 mm height and 130 mm chord length as shown in Fig 14.

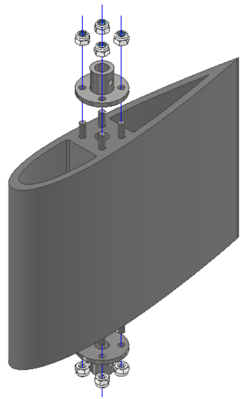

Fig. 14. Rudder modelled in Inventor

The rudder is designed for a maximum angle of −35° to 35° for proper navigation without straining the servomotor past its load limit. A study by [16] concluded that high-lift profiles such as wedge-tail, fishtail and flapped profiles improve the rudder effectiveness but cause additional drag. The symmetrical profile of NACA-0012 compromises between rudder effectiveness and drag. However, to ensure the best mounting procedure, a NACA-0015 profile which has a higher thickness was chosen for the design. The rudder is controlled by a servomotor with electrical input from the surface unit. A fully waterproof housing that encloses the motor was created. The housing includes a shaft coupler to connect the rudder to the servomotor with a straight shaft. The assembly of the 3D design of the servo motor housing assembly is shown in Fig. 15.

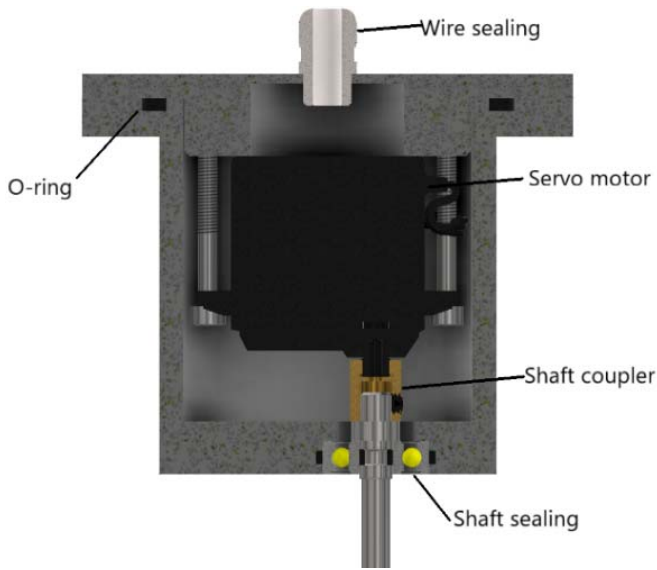

Fig. 15. Cross-section of the 3D design of the servomotor housing

The rudder housing was manufactured from POM material using CNC lathe as shown in Fig. 16. The shaft is manufactured using the same machine from stainless steel. The manufactured servo housing components are shown in Fig. 17.

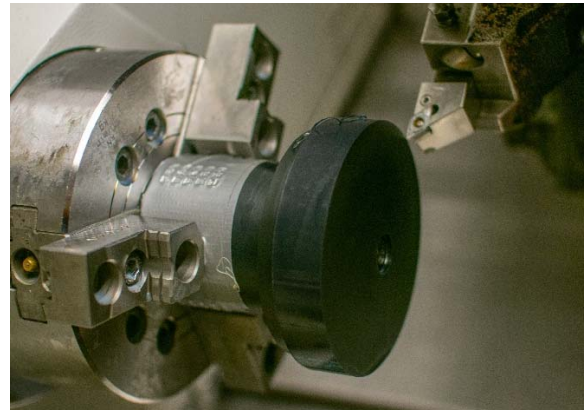

Fig. 16. Servo housing machined by the CNC lathe machine

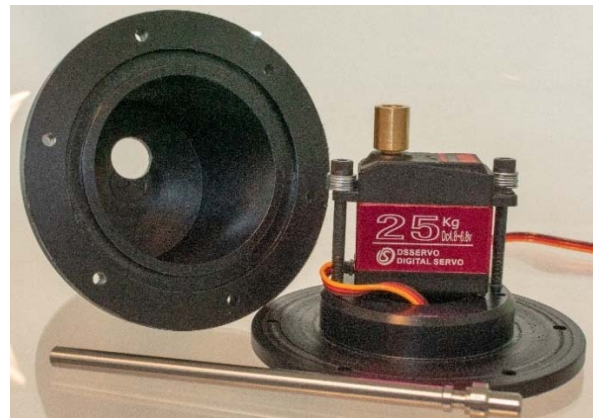

Fig. 17. Servo motor housing components

### E. Tether

A strong connection is required to attach the underwater unit, which lies 5 meters below the surface unit. There are many different types of connections that have been experimented during the design process. The two separate connection points must withstand occurring loads to avoid the glider breaking free from the surface unit. A study by [17] compared the types of tether connections between the sub and the surface unit. In the study, four separate connection types including rigid rod, cable, multi-link, and elastic rod were tested through modeling software. The simulations concluded that there was an insignificant difference in velocity between these connection types while keeping all other variables consistent. Also, there was a damping effect present in the cable and elastic rod. It was determined that the cable connection was most suitable for this application.

The tether is attached to the center of mass of the float and the glider. If the connection were off-center there would be a pitch motion-induced. The cable is attached to the surface unit through the winch assembly. Since the glider has negative buoyancy and the float has positive buoyancy, the cable will always be in tension [2]. Fig. 18 show the manufactured full assembly of the submerged unit.

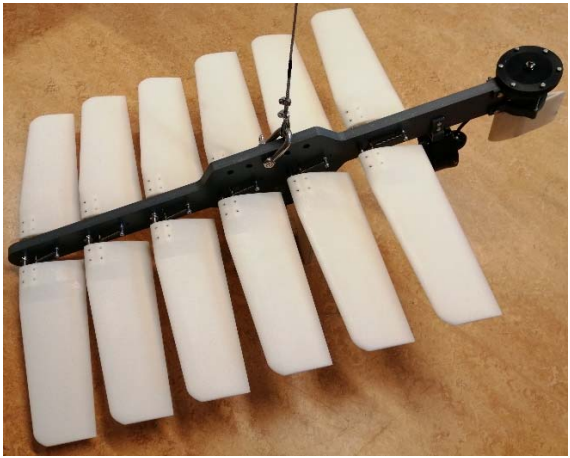

Fig. 18. Assembly of the submerged unit after manufacturing frame, wings, and rudder housing

## IV. ELECTRONICS AND COMMUNICATION

The electronic system provides power supply to all the actuators in the boat, such as the propellers or the servomotors, and is essential in the monitoring of the USV. The main functionalities of the electronics are monitoring the USV, as mentioned; collecting data; steering the boat; communicating with the user, and supplying power to the other units when needed.

### A. Printed Circuit Boards (PCB)

The circuit is controlled by an "ATMEGA2560-16U" microprocessor. A 3D View of the final PCB and a complete circuit schematic can be seen in Fig. 19 and Fig. 20 respectively.

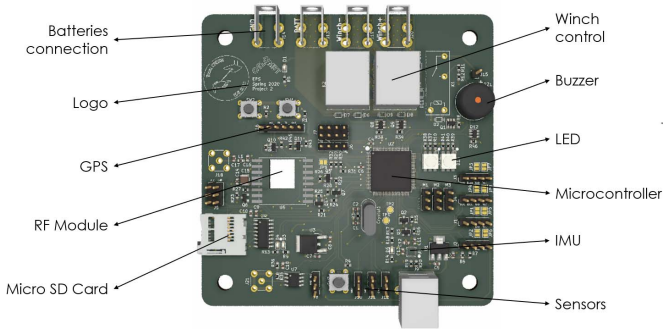

Fig. 19. 3D View of the USV's PCB

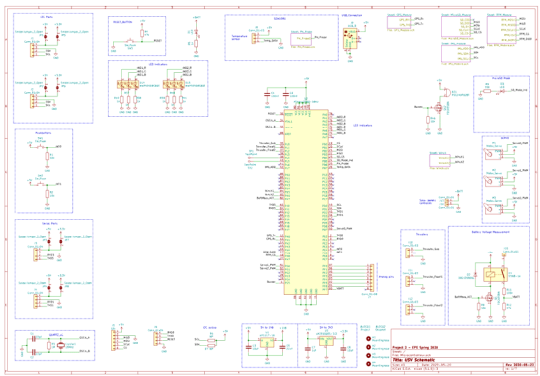

Fig. 20. General circuit schematic from KiCad software

Fig. show the manufactured PCB after soldering all the components and testing all the modules and the blocks on the board.

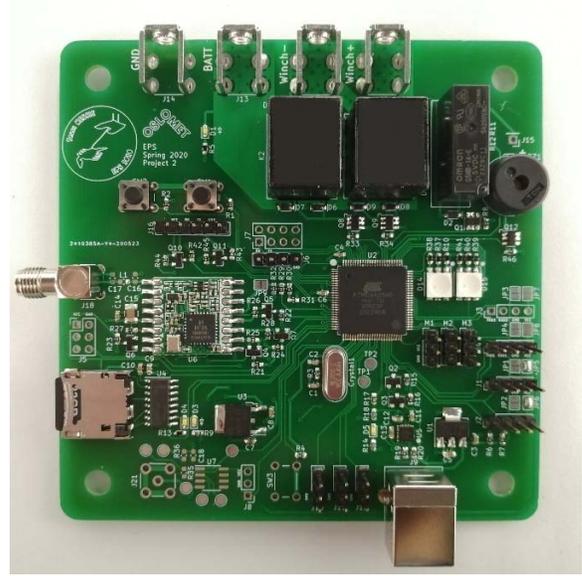

Fig. 21. PCB of the vehicle after manufacturing and soldering

### B. Power supply and management

During long expeditions, electronics need an active power supply which will enable them to keep functioning. Solar panels act as the power supply in this USV. It is not possible to connect the solar panels directly to the PCB since the solar panels' output fluctuates over time depending on the solar irradiance and the electronic components need a stable power supply. For this reason, batteries were installed to accumulate the energy from the panels and, at the same time, provide stable output for the boat's electronics. The USV has a solar panel controller so the batteries can be charged most efficiently, and the highest level of the energy provided by the solar panels can be collected. Overall, the main electrical installation of the USV is composed of solar panels, batteries, a controller, electronic components and actuators and its current flow can be seen in Fig. 22.

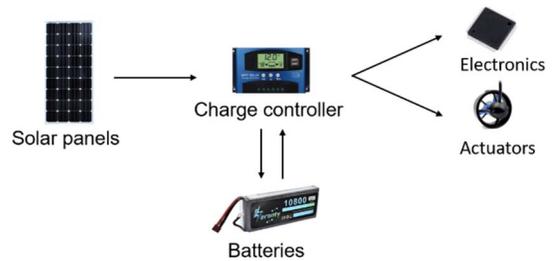

Fig. 22. General power flow of the USV

The USV is powered using 4 lithium polymer (LiPo) batteries connected in parallel, with a total capacity of 43.2Ah and a voltage level of 11.1V, one of them is kept as a backup. This type of batteries has the best performance when used as the power source of solar panels [18]. Li-po batteries have a specific charge and discharge profile which needs to be monitored to maximize their duration.

The solar panels used in the USV have an output of 18 V. The output voltage has been chosen based on the voltage of the batteries and market availability. The solar panel voltage must be higher than the voltage of the batteries to account for losses. The boat's solar installation is composed of three solar panels connected in parallel. As the orientation of the solar panels in the USV is not the same, the solar radiation they receive varies. As such, one of the panels can output a higher voltage than another. In this situation, the solar panels tend to even the voltage by charging the other solar panels. This can lead to burning and malfunctioning of the solar panels. To avoid this situation, diodes can be connected between the solar panels and the controller. The diodes ensure that the current can only flow from the solar panels to the controller and block the other direction. Moreover, the diodes also prevent the battery from trying to charge the solar panels, and therefore, discharge itself.

Two different types of solar controllers exist in the mass market: Maximum Power Point Tracking (MPPT) controllers and Pulse Width Modulation (PWM) controllers. The USV uses an MPPT controller as they have a better conversion rate than the PWM ones and is optimized for LiPo batteries. The solar controller also automatically regulates the charging process of the batteries and provides a current limit for the load at 30 A.

### C. Modules

The USV is equipped with the different electronic modules, which helps in the navigation of the USV, storing the data from the on-board sensors.

#### 1) Inertial Measurement Unit (IMU)

The USV is equipped with a 9-axis IMU, which main functions are to measure USV heading, estimate sea state, and to know whether the surface unit has capsized. The device used in this USV is the "ICM-20948" from "InvenSense Technologies". The IMU uses Microelectromechanical systems (MEMS) technologies for motion monitoring, which reduces the power consumption of the device. The device also incorporates a Digital Motion Processor (DPM) which manipulates the data acquired so it can be interpreted and transmitted. The IMU communicates with the microcontroller using the Inter-Integrated Circuit ($I^2C$) Protocol.

#### 2) Golbal Positioning System (GPS)

A GPS Module is used to estimate the boat's exact position and velocity and aid in its steering system. The model used is the "MTK3339" in an "Adafruit breakout board" which has the possibility of adding an external antenna. It uses the Universal Asynchronous Receiver Transmitter (UART) communication protocol to interact with the microcontroller.

#### 3) Micro Secure Digital (SD) Card

The Micro SD Card Module is an electronic receptacle where a memory card, in this case, a Micro SD Card, can be inserted and interacted with. The module enables the interaction between the microcontroller and the memory card. This bridge allows the microcontroller to write on the card and to read its content. Sending information via wireless communication can be expensive, so only the essential information is transmitted. The other pieces of information, which may include sensors' readings or a more detailed analysis of the USV status, are stored in the memory card. It communicates through the Serial Peripheral Interface (SPI) protocol.

### D. Communication

One of the main features of the USV is the real-time communication between the user and the boat. This functionality is achieved by using a "Raspberry Pi 3", a miniaturized computer that can exchange data with the boat and interact with the user via a web server.

#### 1) Satellite Communication

The satellite communication module used in this USV is the "RockBLOCK" by "Rock7" and the communication with the microcontroller is via UART. Fig. 23 shows the satellite module used in the USV before being mounted to the electronics board.

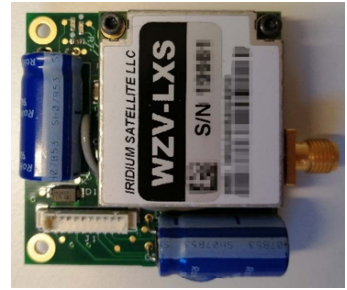

Fig. 23. "RockBLOCK" Module by Rock7

The sensors acquire the data from the ocean and store it in the Micro SD Card in the PCB. Next, the data is sent via the satellite communication module to the RockBLOCK servers, using Iridium satellites. The message is sent as an e-mail to a private address, interpreted by the Raspberry Pi, and stored in a "MySQL" database. The Raspberry Pi uploads the data to an online website that can be consulted by the users. The process can be seen in Fig. 24

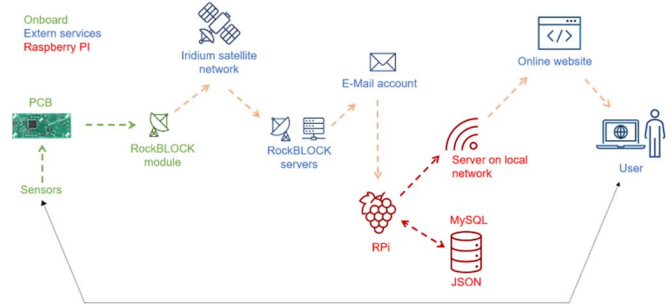

Fig. 24. Block diagram: Sending, receiving and displaying data from the boat

The user can also define new positions for the USV or change the ones that are already programmed. In order to do so, the website has a data input functionality that allows the new positions to be read by the Raspberry Pi and stored in the MySQL database. The Raspberry Pi sends the new data to the PCB through the RockBLOCK servers and module and the Iridium satellites. Finally, the PCB updates or stores the data in the Micro SD Card.

#### 2) Radiofrequency (RF) Communication Module

Radiofrequency is used in short-distance communications as the distance the information can travel depends on the transmitting environment, obstacles, and line of sight. Also, the amount of power transmitted cannot exceed a certain limit. The data can only be received by another module with the same encryption key, ensuring the safety of the information. The module used to establish an RF communication with the user is the "RFM95 breakout" and the communication frequency is 868 MHz. RFM95 module uses a LoRa Radio and SPI. The process for RF communication follows the same path as satellite communication. Although, it uses two RFM95 modules instead of the RockBLOCK modules and servers.

All the modules were tested to check their compatibility with the system and with each other, before they were assembled on the USV, specially that there are three different communication protocols; SPI, UART, and I$^2$C.

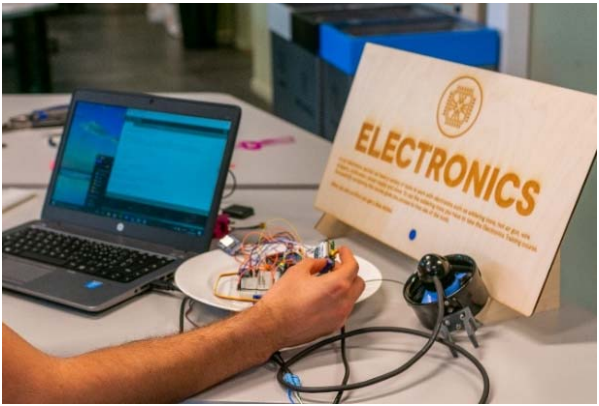

Fig. 25. Testing the different modules with the microprocessor

*3) Web Application*

A web application was created to allow the user to monitor the behavior of the vehicle, get sample readings from the sensors, as well as provide next points on its trajectory. An online website enables the user to control the boat's location, in one-hour intervals, and consult some of the data the USV is collecting. The website displays the latest positions of the vehicle and the sensor readings. Lastly, the user can input new positions for the USV or change the current positions. Fig. 26 and 27 respectively show screenshots from the website for updating the USV trajectory and one of the sensors readings.

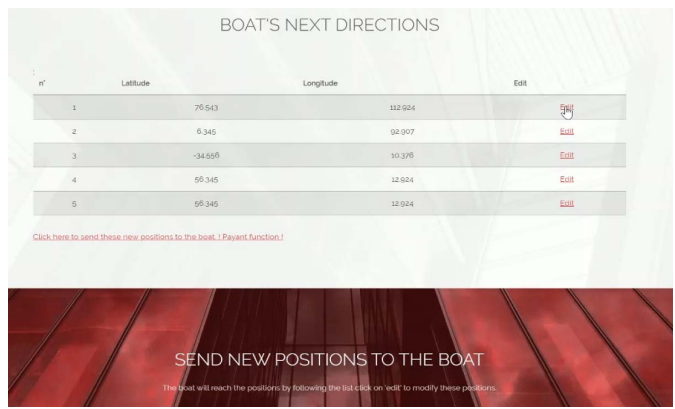

Fig. 26. Updating the trajectory through the website

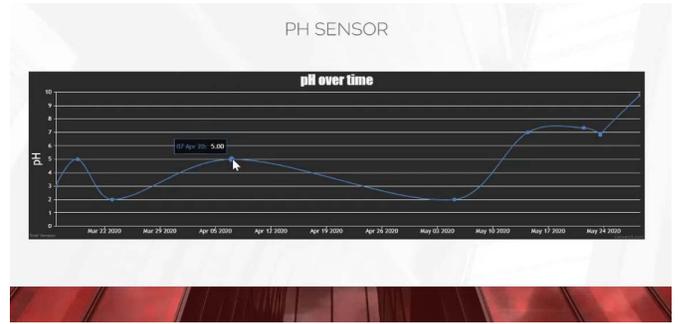

Fig. 27. Monitoring sensors readings from the website

## V. PAYLOAD

The main use of the presented USV is the collection of ocean data. This USV has been equipped with four sensors that measure temperature, pH level, conductivity, and dissolved oxygen of the water. The sensors are mounted on the surface unit, with the possibility of mounting them on a submerged towed unit, since all the sensors are rated to a maximum depth of 60m. the specification of each sensor is mentioned in Table I.

Each sensor is controlled by an interface board designed and fabricated by the manufacturer of the sensors. This board manipulates the signal measured by the sensor to be able to transmit it to the microcontroller. The sensors are manufactured by "Atlas Scientific" a company specialized in underwater environmental sensors.

TABLE I. ON-BOARD SENSORS SPECIFICATIONS.

|  | *Conductivity* | *Temperature* | *Dissolved Oxygen* | *pH* |
|---|---|---|---|---|
| **Range** | 10 μS/cm − 1 S/cm | -200 ℃ − 850 ℃ | 1 − 50 mg/L | 0 − 14 |
| **Accuracy** | +/- 2% | +/- (0.15 + (0.002*t)) | +/- 0.2 mg/L | +/− 0.002 |
| **Response Time** | 90% in 1sec | 90% in 13s | 0.5 mg/L/sec | 95% in 1s |

## VI. SUMMARY AND FUTURE WORK

This paper presented the design of a low-cost wave and solar powered unmanned surface vehicles. The proposed vehicle is small and light enough to be deployed without the need for a crane. The vehicle provides a low-cost platform for ocean monitoring.

Future work includes further developing the design to reduce the weight and dimensions of the USV, as well as providing advance autonomous operation functionalities such as collision avoidance. A cloud service will also be used to store the data from the sensors, in order to avoid the use of any additional servers and hardware onshore. Other sensors will be also added, including a weather station to help provide data not just about the water but also about the weather in the ocean.

Another future development is upgrading the design to be able to deploy and recover miniaturized AUV's, within the scope of OASYS research project in Oslo Metropolitan University [19].

ACKNOWLEDGMENT

This work is within the frame of OASYS research project which is funded by the Research Council of Norway (RCN), the German Federal Ministry of Economic Affairs and Energy (BMWi) and the European Commission under the framework of the ERA-NET Cofund MarTERA.